\documentclass[11pt,a4paper]{article}
\usepackage[hyperref]{naaclhlt2018}
\usepackage{times}
\usepackage{latexsym}
\usepackage{graphicx}
\usepackage{tablefootnote}
\usepackage[utf8]{inputenc}
\usepackage[shortlabels]{enumitem}
\setlist[itemize]{noitemsep,topsep=0pt}

\usepackage{appendix}
\usepackage{url}

\aclfinalcopy % Uncomment this line for the final submission
\def\aclpaperid{930} %  Enter the acl Paper ID here

\newcommand{\argmin}{\mathop{\mathrm{argmin}}}
\newcommand{\argmax}{\mathop{\mathrm{argmax}}}

\title{Community Member Retrieval on Social Media using Textual Information}

\author{\textbf{Aaron Jaech,  Shobhit Hathi,  Mari Ostendorf} \\
  University of Washington \\
  {\tt \{ajaech, shathi, ostendor\}uw.edu}}

\date{}

\begin{document}
\maketitle

\begin{abstract}
This paper addresses the problem of community membership detection using only text features in a scenario where a small number of positive labeled examples defines the community. The solution introduces an unsupervised proxy task for learning user embeddings: user re-identification. Experiments with 16 different communities show that the resulting embeddings are more effective for community membership identification than common unsupervised representations.
\end{abstract}

\section{Introduction}
Active users of social media often like identifying other users with common interests and values. Or, a user may want to find other users that share characteristics with specific accounts that they follow, e.g.\ cartoonists or local food trucks.
Members of such communities of interest are often identifiable via their social network connections, and shared social connections are clearly important in recommendations.
However, shared connections often reflect a subset of a person's interests, and there may be users of interest where any shared connections are distant. In addition, there may be scenarios where there is no explicit social graph, or the full graph is expensive to obtain. In such cases, the language of tweets, blogs, etc.\ is helpful in identifying users with particular interests.

In this paper, we represent users in terms of the text in their communications and introduce a scenario where a user can define a ``community'' by providing a small number of example accounts that are used to train a system for retrieving similar users.
Note that our use of the term ``community'' differs from other online contexts, where members explicitly self-identify with a community (e.g.\ by joining a discussion forum or using a specific hashtag). The community is in the eye of the user issuing the query.

We frame the task of community membership detection as a retrieval problem. A small set of representative accounts selected by the user forms the query, and the system retrieves additional community members from a large index of accounts. The task is loosely related to entity set expansion \cite{pantel2009web}. We make no assumptions about the type of communities that can be handled, and no labeled data is available other than the query. Because the training set (query) is minimal, unsupervised learning is useful for the text representation. We propose the proxy task of person re-identification for learning a user embedding, where the goal is for two embeddings from the same user to be closer to each other than to the embedding of a random user.
The hypothesis is that a representation useful for detecting similarities between posts from the same person made at different times will also do well at identifying similarities between people in the same community. This hypothesis stems from observations that people with shared interests often talk about topics related to these interests, and that they tend to have shared jargon and other similarities in language use \cite{Nguyen2011,DNM2013,Tran2016}.

In this paper, we demonstrate experimentally that the re-identification proxy task is useful with simple models that are suited to the retrieval scenario, and present analyses showing that the approach learns to emphasize words associated with individual interests and polarizing issues.

\section{Model}
The model for community detection
includes: i) a mapping from a user's text (a collection of tweets) to a $k$-dimensional embedding, and ii) a binary classifier for detecting whether a candidate user belongs to the target community. The novel contribution of the work is the proxy re-identification task for learning the user embedding.

\paragraph{User Embedding Model.}

The mapping from text to an embedding could leverage any document-level representation. We focus on a simple weighted bag-of-words neural model for direct comparison to other popular methods, motivated by the fact that many virtual communities form around shared interests in particular topics. Specifically, let $c_{p,i}$ denote the number of times person $p$ uses word $v_i\in V$, where $V$ is the vocabulary, and $w_{p,i}=\log(c_{p,i} +1)$ be the log-scaled word count. Then the user embedding is
\begin{equation}
u_p = \frac{w_p^T \mathbf{E}}{||w_p^T \mathbf{E}||}
\end{equation}
where $w_p=[w_{p,1} \cdots w_{p,|V|}]$ and $\mathbf{E}\in \Re^{|V| \times k}$ is the matrix of word embeddings.

\paragraph{Person Re-identification Learning.}

The embedding matrix $\mathbf{E}$ is learned using a person re-identification objective that encourages embeddings from the same person to be closer than embeddings from different people. We build on the triplet loss function taken from \newcite{schroff2015facenet} used to train a face recognition system. Specifically:
\begin{equation}
\mathbf{E} = \argmin_{\mathbf{E}} \sum_{p_1, p_2 \in \mathcal{P}} \mathrm{cost}(p_1, p_2) ,
\end{equation}
%where
%\begin{equation}
$$
\mathrm{cost}(p_1, p_2) = (1 + \mathrm{d}(u_{p_{1}^{1}},u_{p_{1}^{2}}) - \mathrm{d}(u_{p_{1}^{1}}, u_{p_{2}^{1}}))^+ ,
$$
%\end{equation}
where $\mathrm{d}(x,y)$ is the cosine distance between $x$ and $y$.  $u_{p_{1}^{1}}$ and $u_{p_{1}^{2}}$ are embeddings made from distinct subsets of a single person's Tweets, and $u_{p_{2}^{1}}$ is an embedding made from a subset of another person's Tweets. In practice, we estimate the loss function randomly sampling triplets ($p_1^1$, $p_1^2$, $p_2^1$) from a large training set.

\paragraph{Classifier.}

A logistic regression model with L2 regularization is used for the classifier,
because it is simple but powerful and our scenario has little training data. Simplicity is important because the classifier should be trainable in real-time after receiving the query. The classifier objective is to discriminate the embeddings from the users in the query from a set of user embeddings from the general collection. For the $i$-th user, let $y_i\in\{0, 1\}$ be the binary label indicating whether the user belongs to a particular community and $u_i$ be the user embedding.
The logistic regression model computes the probability that the user belongs to the community according to:
\begin{equation}
    \label{eq:logistic}
    p(y_i = 1|u_i) = \sigma(w^T u_i + b) ,
\end{equation}
where $\sigma(x)=1/(1-e^{-x})$.
During evaluation, the users in the index are ranked according to the maximum log probability ratio
\begin{equation}
    \label{eq:log-ratio}
        \argmax_i\log \frac{p(y_i = 1|u_i)}{p(y_i=0|u_i)} = \argmax_i w^T u_i .
\end{equation}

Because the classifier is linear, we can quickly retrieve the top matching users from the index using approximate nearest-neighbor search
%between $w$ and $u_i$
\cite{kushilevitz2000efficient}. The technique is scalable up to hundreds of millions of users and beyond.

\section{Data}
All data was collected using the Twitter API.\footnote{\href{http://developer.twitter.com/en/docs/api-reference-index}{http://developer.twitter.com/en/docs/api-reference-index}} We used 1,035 randomly selected items from the list of trending topics in the USA during the period April-June 2017 to query for users and collected their most recent 2,000 tweets. % I checked the code to verify that we were sampling from all 2,000 tweets during training
Example trending topics are \#Quantico, RonaldoCristiano, and \#MayDay2017.
(The full list is available with the data.) Each user had at least one Tweet that mentioned a trending topic but their other Tweets could be on any topic.

 We refer to this collection as the ``general population,'' because it was not targeted towards any particular community. In total, we collected around 80,000 such users and used roughly 36,000 for learning user embeddings, 1,000 for learning the community classifiers, and 43,000 for evaluation. The text is mostly in English, but some of it is in Spanish, French, or other languages. A list of the tweet IDs is available.\footnote{\href{https://github.com/ajaech/twittercommunities}{http://github.com/ajaech/twittercommunities}}

To support evaluation with the community detection task, we conducted a second collection (contemporaneous with the first) targeting members that we had identified as belonging to one of 16 communities (Table \ref{table:breakdown}).
To define a ``community,'' volunteers manually selected a set of users that fit with a theme that they had familiarity with.
Thus, the specific 16 communities were determined based on themes of interest to the authors and their friends and colleagues, where we could be reasonably confident about membership decisions. In addition, we tried to avoid themes that might be biased towards well-known celebrities, and we made an effort to have diversity in the characteristics of the communities.
The communities were selected to span a range of topics, sizes (6-130 accounts), individuals vs.\ organizations, and other characteristics. A few of the communities are comprised of organizations rather than individuals such as the high school drama departments and the Pittsburgh food truck communities. (The community names are invented by the authors for purposes of describing the data in this paper; they are not part of the retrieval task.)

The text is lower-cased and some punctuation is removed using regular expressions. Words are formed by splitting on white space. While this strategy will not work for languages that do not delimit words by spaces, these make up a negligible portion of the data.
A 174k vocabulary was created by extracting the unique types that were seen in the tweets  from the general population, as well as selected bigrams extracted using the open source Gensim library using a point-wise mutual information criteria \cite{rehurek_lrec}. The vocabulary included roughly 49k bigrams, 36k usernames and 17k hashtags.
Usernames, hashtags, and URLs are not treated specially and can be part of the vocabulary just like any other word if they occur frequently enough.

\section{Experiments}
\subsection{Experiment Configuration}

The experiments involved comparing different methods of learning user embeddings, all with a weighted bag-of-words modeling assumption:
\begin{itemize}
\item Weighted word2vec (W2V) using default\footnote{The default configuration uses
a window of $\pm7$ words. We also tried using a window of 50 words, which roughly matches the context used in other methods,
but community detection performance was significantly worse.}
skip-gram training \cite{mikolov2013distributed};
\item Latent Dirichlet allocation (LDA) \cite{Blei+03}, using default settings from the  Scikit Learn library \cite{scikit-learn};
\item Person re-identification with random initialization (RE-ID); and
\item Person re-identification with W2V initialization (RE-ID, W2V init).
\end{itemize}
Both count-weighted W2V and LDA have been used as unsupervised representations in Twitter classification tasks, as noted in Section~\ref{sec:related}.
Default configurations are used because there is insufficient data to have a separate validation set.

For all methods, the same vocabulary, final dimension (128), unit vector normalization strategy, and logistic regression model training were used. The embeddings are trained on the 36k user general data, randomly sampling pairs of users $p_1$ and $p_2$ and then sampling 50 tweets at a time without replacement to create $u_{p_{1}^{1}}$, $u_{p_{1}^{2}}$, and  $u_{p_{2}^{1}}$.
The logistic regression models are trained on the 1K user general training pool, using the 50 most recent tweets for each user.
Because there are so few labeled examples for most communities, training and evaluation is done using a leave-one-out strategy with the positive samples but including all of the 1K negative samples. For each of the $N$ classifiers (corresponding to $N$ labeled samples), the test set is the left-out positive example and the 43K general user test pool.
Also because of training limitations, there is no tuning of the regularization weight; the default weight of 1.0 is used. Tuning may be useful given a collection of training and testing communities.
Performance is averaged over the $N$ classifiers (corresponding to the $N$ labeled samples). Two evaluation criteria are used: a retrieval metric (inverse mean reciprocal rank or 1/MRR) \cite{voorhees1999trec} and a detection metric (area under the curve or AUC).

\subsection{Results}

Table \ref{table:stats} shows retrieval results averaged across all communities. The RE-ID model outperforms the W2V and LDA baselines for both criteria, with substantial gains in 1/MRR (lower is better).
Further, the version of RE-ID initialized with word2vec did better than the one that was initialized randomly even though the randomly initialized version was trained for twice as long.

\begin{table}[ht]
\centering
\begin{tabular}{lrr}
\textbf{Strategy} & \textbf{AUC} & \textbf{1/MRR} \\ \hline
W2V & 93.9        & 846 \\
LDA &  95.0 & 501 \\
RE-ID (rand. init)       & 98.0         & 24 \\
RE-ID (W2V init)      & \textbf{98.5}       & \textbf{12}  \\
\end{tabular}
\caption{Performance of different model variants.}
%on the community detection task.}
\label{table:stats}
\end{table}

A breakdown of the best model performance by community is given in  Table \ref{table:breakdown}.
Sample size does not seem to be a good indicator of performance: the two smallest communities (Cartoonists, Fresno City Council) had the worst and one of the best results, respectively. Anecdotally, we observed that the sample of cartoonists were more likely to Tweet about topics outside their main interest (e.g., politics or sports). We hypothesize that the diversity of interests of the members of a community affects the difficulty of the retrieval task, but our test set is too small to confirm this hypothesis.

\begin{table}[ht]
\centering
\begin{tabular}{lrr}
\textbf{Community}  & \textbf{Size}   & \textbf{1/MRR} \\ \hline
Cartoonists              & 8  & 58.1  \\
Chess Stars              & 14 & 5.4 \\
Conan Show Writers       & 12  & 4.7 \\
Fashion Commentators     & 11 & 8.3 \\
Fresno City Council      & 6 & 3.0 \\
Hedge Fund Managers      & 11 & 25.7 \\
H.S Drama Departments    & 18 & 2.3 \\
Mathematicians           & 11 & 32.6 \\
NLP Researchers          & 50  & 4.9 \\
Pittsburgh Food Trucks   & 15  & 3.3 \\
Police Dogs              & 16  & 2.7 \\
Professional Economists  & 11 & 3.6 \\
SCOTUS Reporters\tablefootnote{People who write news articles about the Supreme Court of the United States.}         & 16  & 1.9 \\
The Stranger Reporters\tablefootnote{The Stranger is a small weekly newspaper.}   & 11 & 8.3 \\
Ultimate Frisbee Players & 130 & 6.7 \\
Ultramarathon Runners    & 28 & 14.6
\end{tabular}
\caption{W2V+RE-ID results by community}
\label{table:breakdown}
\end{table}

These results may underestimate performance, because there is a chance that some users in the general population test data may actually belong to one or more of our test communities, i.e. there could be mislabeled data. To assess the potential impact, we manually checked the top ten false positives for each community for mislabeled users. We did discover some mislabeled examples for the economist, hedge fund manager, and ultramarathon runner communities. For the most part, the top ranked users from the general population tended to be people from related communities. For example, the top false ultimate frisbee users contained people who wrote about their participation in tournaments for other sports such as soccer.

\subsection{Analysis}

The finding that the W2V-initialized RE-ID model is significantly better than W2V raises the question: how do the embeddings learned by the re-identification task differ from the ones learned by the word2vec objective? To investigate this, we looked at the 1,000 words in the \mbox{RE-ID} model with embeddings that were farthest (in Euclidean distance) from its word2vec initialization. These top words disproportionately contain Twitter user handles, so some social network structure is captured. Using agglomerative clustering, we found groups of words that centered around frequent words used in particular regions (foreign words, dialects) or cultures (sociolects), associated with hobbies or interests (specific sports, music genres, gaming), or polarizing topics (political parties, controversial issues). At least one of the top tokens was the username of an account later identified as being sponsored by the Russian government to spread propaganda during the United States presidential election, e.g., ``ten\_gop'' in Table~\ref{table:changed_words} of the Appendix.

We also looked at which communities are closest in the embedding space. We
represent a community with the average of the member embeddings and use a normalized cosine distance for similarity.
The two nearest neighbors are Mathematicians and NLP researchers, which are also close to the next two nearest neighbors, Hedge Fund Managers and Professional Economists.

To interpret what the model as a whole captured, we found the top scoring tweets for each held-out user (creating an embedding for a single tweet) according to the logistic regression model.
Representative examples include ``recurrent  neural\_network grammars simplified and analyzed'' for NLP Researchers, and ``we're looking\_forward to seeing you opening\_night may 24th love the cast of high\_school musical'' for High School Drama clubs. Examples for additional communities are included in the appendix.
The results provide insight into the community member identification decision.

\section{Related Work}
\label{sec:related}
One notion of community detection involves discovering different communities within a collection of users \cite{Chen2009LocalCI,Di2011,fanitemporally}. A related task is making recommendations of friends or people to follow \cite{gupta2013wtf,Yu2016}.  In contrast, our task involves identifying other members of a community, which is specified in terms of a set of example users.  These tasks use different learning frameworks (our work uses supervised learning), but the features (social network and/or text cues) are relevant across tasks.
Our task is perhaps more similar to using social media text to predict author characteristics such as personality \cite{golbeck2011predicting}, gang membership \cite{wijeratne2016word}, geolocation \cite{han2014text}, political affiliation \cite{makazhanov2014predicting}, occupational class \cite{preoctiuc2015analysis}, and more. Again, a commonality across tasks is the frequent use of unsupervised representations of textual features.

In representing text, a common assumption is that community language reflects topical interests, so representations aimed at topic modeling have been used, including LDA \cite{Pennacchiotti2011AML} and tf-idf weighted word2vec embeddings \cite{Boom2016RepresentationLF, wijeratne2016word}. \newcite{Yu2016} compute a user embedding by averaging tweet embeddings.
Other work investigates methods for learning embeddings that integrate text and social network (graph or text-based) features \cite{benton2016learning}.

The work closest to ours is by \newcite{fanitemporally}, which learns embeddings that are close for like-minded users, where like-minded pairs are identified by a deterministic algorithm that leverages timing of related posts. Our approach requires no additional heuristics for defining user similarity, but instead relies on an objective that maximizes self-similarity and minimizes similarity to other users randomly sampled from a large general pool.

Our person re-identification proxy task makes use of the triplet loss used to learn person embeddings for face recognition \cite{schroff2015facenet}. In image processing, person re-identification refers to the task of tracking people who have left the field of view of one camera and are later seen by another camera \cite{Bedagkar14}. It is different from our proxy task and the methods are not the same.

\section{Conclusion}
In summary, this paper defines a task of community member retrieval based on their tweets, introduces a person re-identification task to allow community definition with a small number of examples, and shows that that the method gives very good results compared to word2vec and LDA baselines. Analyses show that the user embeddings learned efficiently represent user interests. The text embeddings are largely complementary to the social network features used in other studies, so performance gains can be expected from feature combination.

While our experiments use a bag-of-words representation, as in most related work, the re-identification training objective proposed here can easily be used with other methods for deriving document embeddings, e.g.\ \cite{Le2014ICML,Kim2014}.

\section*{Acknowledgements}
The authors thank the anonymous reviewers for their feedback and helpful suggestions.

\bibliography{mybib}
\bibliographystyle{acl_natbib}

\section*{Appendix}
%\documentclass[11pt,a4paper]{article}
%\usepackage[hyperref]{naaclhlt2018}
%\usepackage{times}
%\usepackage{latexsym}
%\usepackage{graphicx}
%\usepackage{tablefootnote}
%\usepackage[utf8]{inputenc}
%\usepackage[shortlabels]{enumitem}
%\setlist[itemize]{noitemsep,topsep=0pt}

%\begin{document}

Supplementary materials include examples of representative tweets for each community (Table~\ref{table:more_top_tweets}) and lists of the words that change the most between Word2Vec and the person re-identification task (Table~\ref{table:changed_words}).

\begin{table*}[ht]
\centering
\begin{tabular}{lp{11.5cm}} 
\textbf{Community}                  & \textbf{Selected Tweet}   \\ \hline
Chess Stars                & @chesscom yep karpov well\_done twittersphere  \\
Professional Economists    & \#china real\_estate as long as liquidity remains ample this will continue            \\
Fashion Commentators       & rihanna's fenty corp creative\_director jahleel weaver styles the collection on 3 muses \\
Fresno City Council        & gr8 resource developed by our local @citdfresno on how to export @cityoffresno @fresnocountyedc lee\_ann eager \\
High School Drama          & we're looking\_forward to seeing you opening\_night may 24th love the cast of high\_school musical  \\
Mathematicians             & forms of knowledge of advanced mathematics for teaching (i wrote a thing ) \\
NLP Researchers            & recurrent neural\_network grammars simplified and analyzed  \\
Police Dogs                & when a trained police dog is placed with another handler they complete a re handling course to be licensed normally 2\_weeks             \\
SCOTUS Reporters           & as supreme\_court throws out two gop-drawn congressional\_districts as unconstitutional racial gerrymanders \\
Ultramarathon Runners      & we're covering the lake sonoma 50 mile live on saturday tell your friends spread the word and get ready \\  
\end{tabular}%
\caption{Top tweets for selected communities. Underscore is used to join bigrams.}
\label{table:more_top_tweets}
\end{table*}

\begin{table*}[ht]
\centering
\begin{tabular}{lp{12.7cm}}
\textbf{Interpretation} & \textbf{Top Words}             \\ \hline
Languages \&   &  \textbullet~à, ça, j'ai, quand, c'est, avec, sur, dans\_le \\
Dialects       &  \textbullet~é, não, melhor, tem, mesmo, só, mais, hoje, uma, tá, já \\
               &  \textbullet~es\_un, más, jugar, en\_el, maduro, jajajaja\\ 
               &  \textbullet~bruh, dawg, @iamakademiks, black\_women, @chancetherapper, lmaooo, y'all, tryna \\ \hline
Sports         &  \textbullet~@mlb, baseball, bullpen, @angels, mets, mlb \\
               &  \textbullet~arsenal, mate, liverpool, @manutd, mourinho, \#mufc \\ 
               &  \textbullet~@nhl, hockey, nhl, leafs, @nhlblackhawks, @nhlonnbcsports \\
               &  \textbullet~xd, @playoverwatch, \#ps4share, anime, @keemstar, overwatch, twitch, @nintendoamerica, gaming \\ \hline
Music          &  \textbullet~@niallofficial, @harry\_styles, @louis\_tomlinson, @ashton5sos, @shawnmendes, @ethandolan, @graysondolan, @michael5sos, @danisnotonfire \\ \hline
Political      & \textbullet~@indivisibleteam \#resist, \#trumpcare, @ezlusztig, @kurteichenwald, @georgetakei, @sarahkendzior, @repadamschiff, @malcolmnance, @lawrence \\ 
               & \textbullet~@mitchellvii, @prisonplanet, @realjameswoods, @jackposobiec, @bfraser747, @cernovich, @ten\_gop, \#maga\\ \hline
Other       & \textbullet~tories, labour, corbyn, \#auspol, tory, mum, nhs, lads scotland \\

\end{tabular}
\caption{Clusters of words which change the most between Word2Vec and the person re-identification task. }
\label{table:changed_words}
\end{table*}

%\end{document}

\end{document}